\documentclass{article}

\usepackage{microtype}
\usepackage{graphicx}
\usepackage{subfigure}
\usepackage{booktabs} 

\usepackage{hyperref}

\usepackage[accepted]{icml2024}

\usepackage{amsmath}
\usepackage{amssymb}
\usepackage{mathtools}
\usepackage{amsthm}

\usepackage{graphicx}
\usepackage{booktabs}
\usepackage{multirow}

\usepackage[capitalize,noabbrev]{cleveref}

\theoremstyle{plain}

\theoremstyle{definition}

\theoremstyle{remark}

\usepackage[textsize=tiny]{todonotes}

\icmltitlerunning{Discrete Latent Perspective Learning for Segmentation and Detection}

\begin{document}

\twocolumn[
\icmltitle{Discrete Latent Perspective Learning for Segmentation and Detection}



\icmlsetsymbol{equal}{*}

\begin{icmlauthorlist}
\icmlauthor{Deyi Ji}{ustc,alibaba}
\icmlauthor{Feng Zhao}{ustc}
\icmlauthor{Lanyun Zhu}{sutd}
\icmlauthor{Wenwei Jin}{alibaba}
\icmlauthor{Hongtao Lu}{sjtu}
\icmlauthor{Jieping Ye}{alibaba}

\end{icmlauthorlist}

\icmlaffiliation{ustc}{University of Science and Technology of China}
\icmlaffiliation{alibaba}{Alibaba Group}
\icmlaffiliation{sutd}{Singapore University of Technology and Design}
\icmlaffiliation{sjtu}{Dept. of CSE, MOE Key Lab of Artificial Intelligence, AI Institute, Shanghai Jiao Tong University}

\icmlcorrespondingauthor{Feng Zhao}{fzhao956@ustc.edu.cn}
\icmlcorrespondingauthor{Jieping Ye}{yejieping.ye@alibaba-inc.com}

\icmlkeywords{Machine Learning, ICML}

\vskip 0.3in
]



\printAffiliationsAndNotice{}  

\begin{abstract}
In this paper, we address the challenge of Perspective-Invariant Learning in machine learning and computer vision, which involves enabling a network to understand images from varying perspectives to achieve consistent semantic interpretation. While standard approaches rely on the labor-intensive collection of multi-view images or limited data augmentation techniques, we propose a novel framework, Discrete Latent Perspective Learning (DLPL), for latent multi-perspective fusion learning using conventional single-view images. DLPL comprises three main modules: Perspective Discrete Decomposition (PDD), Perspective Homography Transformation (PHT), and Perspective Invariant Attention (PIA), which work together to discretize visual features, transform perspectives, and fuse multi-perspective semantic information, respectively. DLPL is a universal perspective learning framework applicable to a variety of scenarios and vision tasks. Extensive experiments demonstrate that DLPL significantly enhances the network's capacity to depict images across diverse scenarios (daily photos, UAV, auto-driving) and tasks (detection, segmentation).
\end{abstract}

\section{Introduction}
\label{sec:intro}

\begin{figure}
    \centering
    \includegraphics[width=1\linewidth]{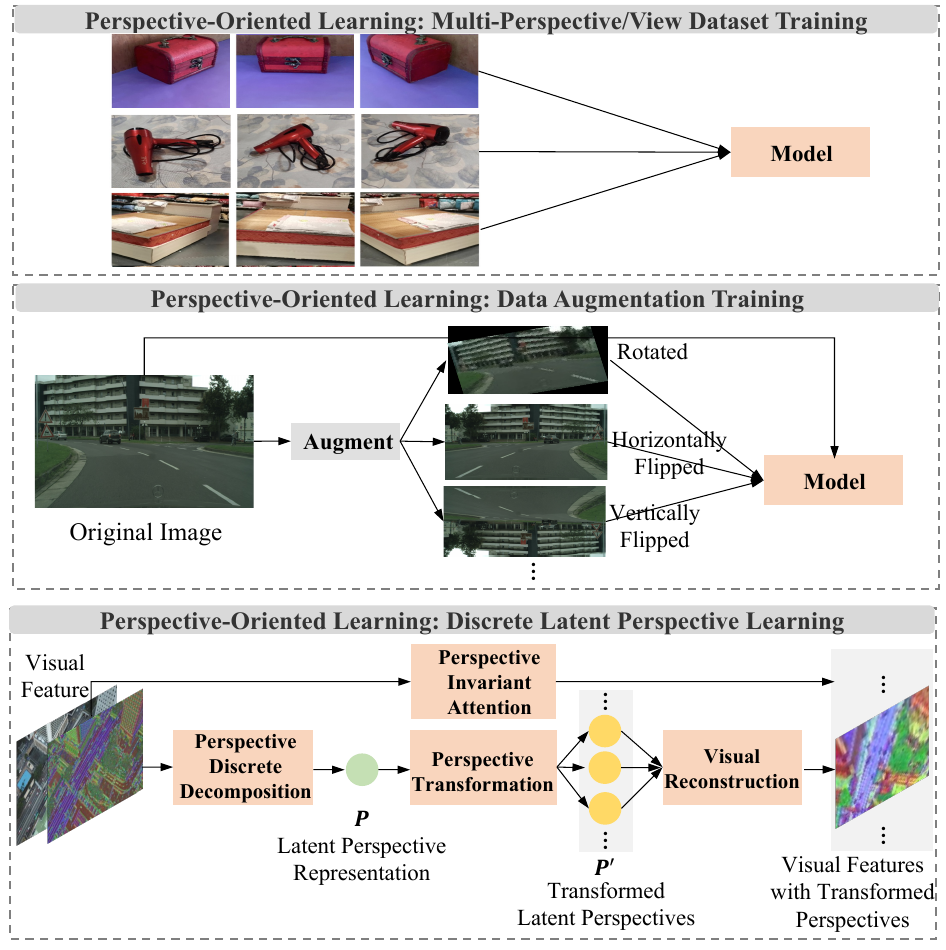}
    \caption{The comparison of perspective-oriented learning methods: Perspective Augmentation, Multi-View Data Training, and the proposed Discrete Latent Perspective Learning. 
    }
    
    \label{fig_intro1}
\end{figure}

According to the theory of photography \cite{photo_theory, bengio2013representation}, image perspective/viewpoint refers to the perception of depth and spatial relationships between objects within an image. In the fields of machine learning and computer vision, current research often employs a variety of methods to enhance networks' \textbf{Perspective-Invariant Learning} \cite{YU2018109,LO1997383}, which involves enabling the network to learn scene information from different viewpoints and ensuring consistency in semantic learning across these varying perspectives, thus improving the portrayal of semantic information within images\footnote{In this paper, the terms ``perspective" and ``viewpoint" are used interchangeably, the former is derived from the theory of photography, while the latter is more commonly employed in general discourse.}. 
As shown in Figure \ref{fig_intro1}, the most commonly used approaches include: (1) collecting multi-view images of a given scene to conduct multi-view training \cite{mvimgnet}, and (2)  utilizing data augmentation techniques \cite{autoaugment}, such as rotation and flipping, to increase the diversity of perspectives.

\begin{figure}
    \centering
    \includegraphics[width=1\linewidth]{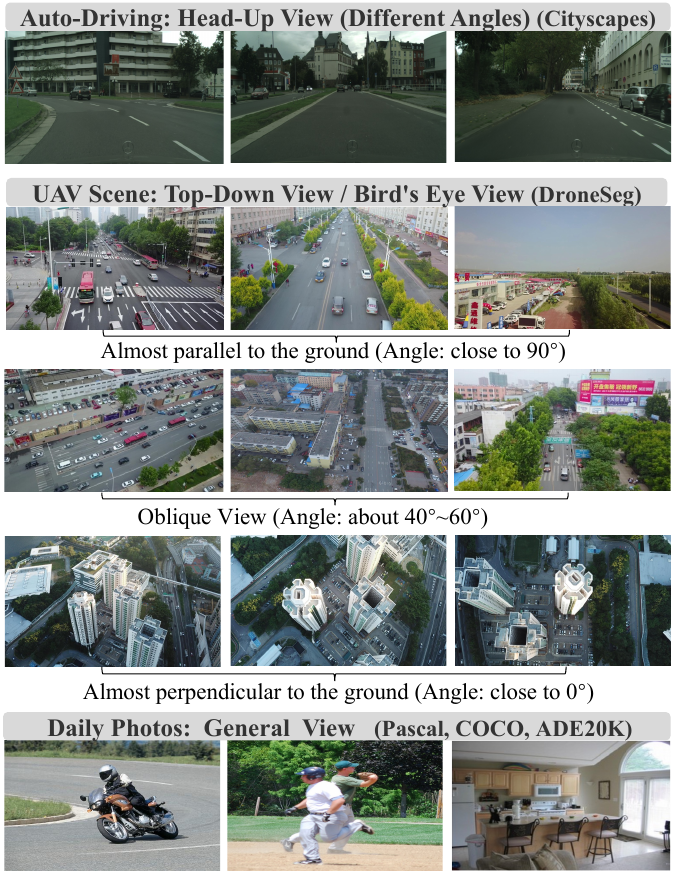}
    \caption{The variation in image perspectives in three typical scenarios: auto-driving, UAV flight, and daily photos. UAV images exhibit more diverse and dynamic viewpoints due to variations in flight heights, angles, and other factors. However, these datasets lack multi-view images for a single scene, preventing existing networks from effective perspective-invariant learning of the images. 
    }
    
    \label{fig_intro}
\end{figure}

However, in the former case, the acquisition of multi-view images is significantly more challenging than that of single-view images, since the latter can rely on fixed cameras with a wide range of sources (e.g., Internet, vehicle-mounted cameras, surveillance cameras), whereas multi-view images are heavily dependent on manual collection, requiring multi-directional shooting of a scene, which is both time-consuming and labor-intensive. In practice, the costs associated with this method are difficult to justify, and the organization and annotation of multi-view images are also more expensive due to the need to ensure consistency in labeling. In the latter case, data augmentation methods often employ hand-crafted rotation operations to augment image perspectives, aiming to enhance training diversity. However, these operations yield limited performance improvements as they generate similar and rigid results, failing to capture the practical perspective changes encountered in real-world applications, such as Unmanned Aerial Vehicle (UAV) flights, auto-driving and daily photos, as shown in Figure \ref{fig_intro}.

In this paper, we propose to perform latent multi-perspective fusion of Invariant Learning within the discrete feature space, based on conventional single-view images. To achieve this, we present a  systematic Discrete Latent Perspective Learning (DLPL) framework, comprising Perspective Representation, Transformation, and Multi-Perspective Fusion. 
Firstly, since perspective lacks a clear formulation in current research, we begin with  Perspective Discrete Decomposition (PDD) to discretize the visual feature\footnote{In this paper, we define ``visual feature" as characteristics extracted directly from input images using conventional CNNs or ViTs, primarily containing visual information and representing the image's semantics. This aligns with the understanding in existing works \cite{caron2020unsupervised,caron2018deep,jing2020self}. ``Perspective feature/representation" refers to features related to the image's perspective, extracted from visual features using our designed Perspective Discrete Decomposition (PDD)  module, mainly representing the image's perspective information. Additionally, visual features can be reconstructed from perspective representations through Visual Reconstruction (VR) module.} to decompose the latent perspective representation. Following this, in line with photography theory, we treat the perspective as a spatial projection of the image, and transform it to generate new descriptive perspectives with the Perspective Homography Transformation (PHT) module. These new perspectives correspond to alternatives spatial projection of the image. Finally, we integrate both the perspectives for fusion learning, with Perspective Invariant Attention (PIA), to obtain perspective-invariant semantic information of the image, ultimately enhancing the model's ability to depict the image. It is important to note that DLPL is a universal learning framework for image perspectives and applicable broadly to different scenarios (daily photos, UAVs, Auto-driving) and various vision tasks (classification, detection, segmentation).

The overall contributions can be summarized as follows: 

\begin{itemize}
    \item DLPL is a general perspective-invariant learning framework that enables multi-perspective fusion learning within latent feature space for single-view images.
    
    \item Specifically, DLPL systematically includes key processes such as Perspective Representation (PDD), Perspective Transformation (PHT), and Multi-Perspective fusion (Multi-Level PIA).
    
    \item Experiments demonstrate that DLPL achieves impressive results across a wide range of scenarios (daily photos, UAV, Auto-driving) and different vision tasks (detection and segmentation).
\end{itemize}

\begin{figure*}
    \centering
    \includegraphics[width=0.89\linewidth]{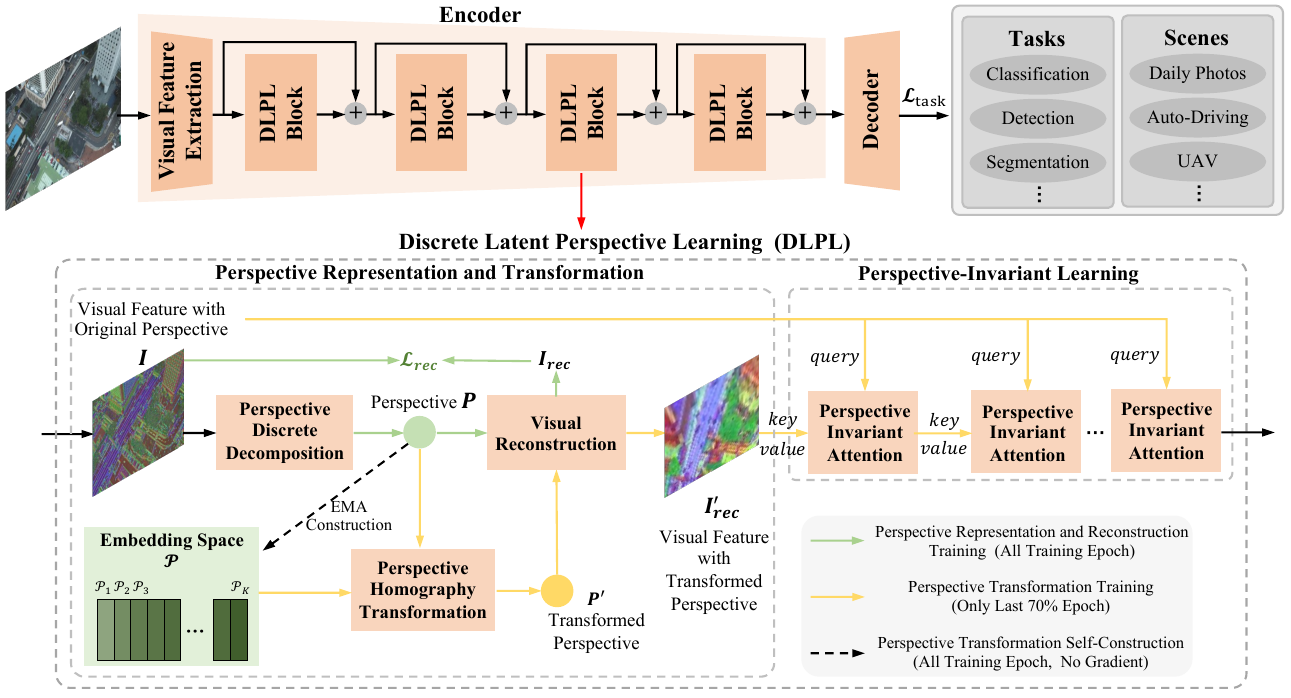}
    \caption{The architecture of DLPL framework within a Vision Transformer (ViT) setup, which is divided into an encoder with a Visual Feature Extraction module and four DLPL blocks, and a task-specific decoder. The DLPL block include: (1) Perspective Discrete Decomposition (PDD) module decomposes input visual features $\mathbf{I}$ into a latent representation $\mathbf{P}$; (2) Visual Reconstruction (VR) module reconstructs visual features from $\mathbf{P}$ with a loss function $\mathcal{L}_{rec}$; (3)
    Perspective Embedding Space $\mathcal{P}$ constructed through EMA, based on $\mathbf{P}$; (4) Perspective Transformation leverages Homography for perspective variation, followed by VR to obtain reconstructed visual features $\mathbf{I}'_{rec}$ of the transformed perspective $\mathbf{P}'$. Finally, Perspective-Invariant Learning fuses the features of the original and transformed perspectives using Perspective-Invariant Attention (PIA) to learn semantic information invariant to perspective changes. 
    }
    \label{fig_method}
\end{figure*}

\section{Related Work}

\subsection{Scene Understanding}

In existing visual datasets, changes in the perspective of images are very common and pose numerous challenges to image understanding. Among these, high-altitude scenes, such as those captured by UAVs or satellites, feature particularly rich variations in viewpoints. Initial works on core high-altitude vision tasks like object detection and semantic segmentation \cite{fu2022panoptic, urur, gpwformer,chen2024reasoning3d} often adapt network architectures from methods designed for natural scenes to establish baseline results. In contrast, the variations in perspectives within auto-driving \cite{cityscapes} and daily photo \cite{ade20k} scenarios are not as pronounced. A plethora of methods have been proposed, including early CNN-based approaches \cite{deeplabv3+, ocrnet, stlnet, cagcn, sstkd, cdgc, changenet, homview, wang2021learning, ipgn,wang2023fvp,feng2018challenges, mscnn} and recent methodologies based on Vision Transformers (ViTs) \cite{segformer, setr, poolformer, detr, zhu2023continual, zhu2023gabor, zhu2023add}, as well as emerging studies on large vision models and multimodal models \cite{sam, llafs, chen2023sam, ibdnet}. Nevertheless, economically and effectively addressing the challenges posed by perspective changes remains a difficult issue to be resolved.

\subsection{Perspective-Oriented Learning} 

Current work on Perspective Learning primarily focuses on data augmentation techniques based on single-view images \cite{cityscapes, ade20k} and joint training with multi-view datasets \cite{mvimgnet,chen2023deep3dsketch}. A substantial body of research on data augmentation techniques \cite{MUMUNI2022100258, autoaugment,chen2023reality3dsketch} has been conducted in existing works, including earlier strategies based on manual rules and later studies that are based on trained learning approaches. The construction of multi-view datasets is a more recent area of research, which involves capturing scenes and targets from multiple orientations. The creation of such datasets often demands significant human effort, and in tasks requiring fine-grained annotation like segmentation, there is the challenge of maintaining consistency across multiple orientations, that is, ensuring that the fine-grained annotations for the same target from different perspectives have consistent semantic meanings.

\begin{figure*}
\centering
  \includegraphics[width=1\linewidth]{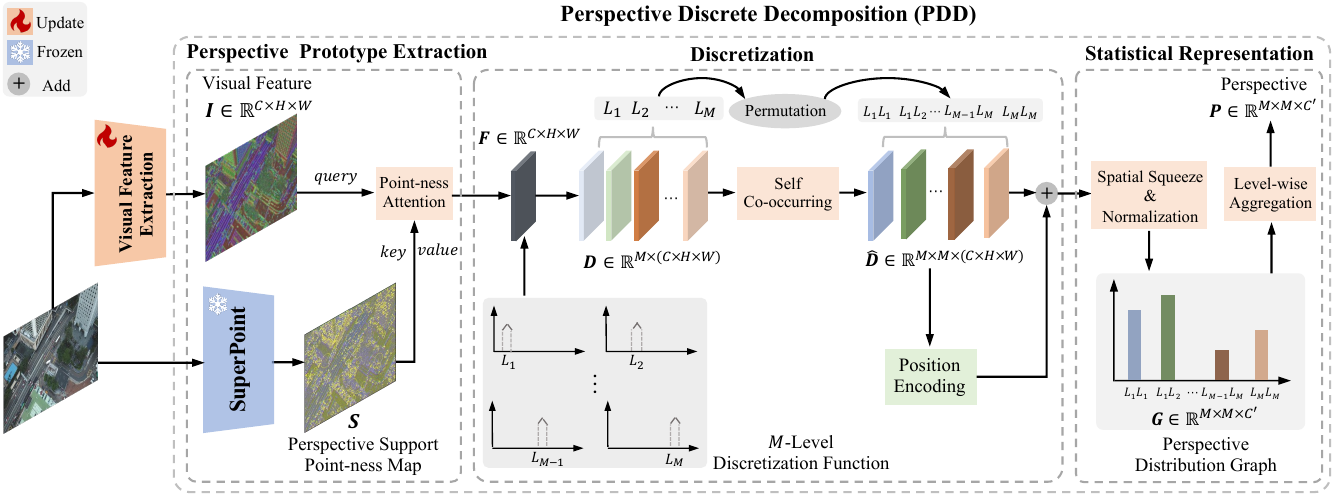}
  \caption{The illustration of PDD. It takes visual features $\mathbf{I}$ as input and processes through three key steps: (1) Extraction of perspective prototypes from a point-ness map $\mathbf{S}$ using spatially structured supporting points; (2) Discretization of the continuous perspective distribution into $M$ levels within the latent feature space; (3) Construction of a statistical perspective graph $\mathbf{G}$ by encoding the intensity and positional relationships of discretized levels. This process yields the latent perspective feature map $\mathbf{P}$.}
  \label{fig_pdd}
\end{figure*}

\section{ Discrete Latent Perspective Learning} \label{sec_method}

\subsection{Top-Down Overview} \label{sec_overview}

DLPL is a flexible and systematic framework for enhancing perspective-invariant learning capabilities in both CNNs and ViT networks. Specifically, as shown in Figure \ref{fig_method}, taking a general ViT  architecture as an example, DLPL can be implemented in the form of Transformer Blocks. The encoder includes a basic Visual Feature Extraction module and four DLPL Blocks. The decoder, on the other hand, chooses the appropriate structure based on the specific task type (classification/segmentation/detection).

\noindent\textbf{Extracting Original Perspective Representation (Section \ref{sec_pdd}).} Within each DLPL Block, given the input visual feature $\mathbf{I}$ (which contains the original image perspective), we extract the original perspective representation from $\mathbf{I}$, by the Perspective Discrete Decomposition (PDD) module.

\paragraph{Reconstructing Visual Feature (Section \ref{sec_vr}).} In order to ensure that the visual feature can be reconstructed from the perspective representation, we design a Visual Reconstruction (VR) module. We constrain $\mathbf{I}$ and the reconstructed visual feature $\mathbf{I}_{rec}$ to be equivalent through a loss function $\mathcal{L}_{rec}$. This is a self-training process.

\paragraph{Perspective Space Construction (Section \ref{sec_ema}).} Concurrently, during training, based on the perspective $\mathbf{P}$ of each image, we can incrementally build the Perspective Embedding Space $\mathcal{P}$ for the entire dataset. To maintain training stability, we update $\mathcal{P}$ using an Exponential Moving Average (EMA) method without backpropagating gradients.

\paragraph{Perspective Transformation (Section \ref{sec_pht}).} As the VR's reconstruction ability and $\mathcal{P}$ become stable, we leverage the spatial Homography properties of image perspective, combined with the distribution of $\mathcal{P}$, to transform $\mathbf{P}$ and obtain another perspective $\mathbf{P}'$, of the image. Then, through VR, we obtain the corresponding visual feature $\mathbf{I}'_{rec}$ of $\mathbf{P}'$.

\paragraph{Perspective-Invariant Learning  (Section \ref{sec_pia}).} We fuse the visual feature $\mathbf{I}$ (with original perspective $\mathbf{P}$) and  $\mathbf{I}'_{rec}$ (with transformed perspective $\mathbf{P}'$) using multi-level Perspective-Invariant Attention (PIA) to learn the network's semantic expression of image information under different perspectives.

In training, Visual Reconstruction and Perspective Space Construction accompany the entire training process. In contrast, Perspective Transformation training begins after the first two processes stabilize, at the 30\% of the entire training epoch. Perspective-Invariant Learning also occurs throughout the entire training process. During the initial 30\% training epoch, since $\mathbf{P}'$ is not ready, PIA degenerates to Self-Attention learning solely based on $\mathbf{P}$.

\subsection{Perspective Discrete Decomposition} \label{sec_pdd}

Since perspective  lacks a clear formulation in current research, we begin with the PDD module. Drawing from visual features extracted by the previous module, we construct prototypes of image perspectives based on key supporting vectors of the perspective. These are then discretized within the latent feature space and statistically analyzed to decompose the implicit perspective distribution map $\mathbf{P}$, as shown in Figure \ref{fig_pdd}.

Formally, PDD receives the visual feature $\mathbf{I}^{C\times H \times W}$ as input and produces a perspective feature $\mathbf{P} \in \mathbb{R}^{M\times M \times C'}$, where $C, C'$ are the feature channels, $H, W$ are the spatial dimensions, $M$ is the discretization levels.  Generally, PDD exploits the explicit well-formulated distribution of structured supporting points which build up the spatial structured prototype to tell the implicit perspective information. In concrete, PDD consists  of three procedures:

\begin{figure}
  \centering
  \includegraphics[width=1\linewidth]{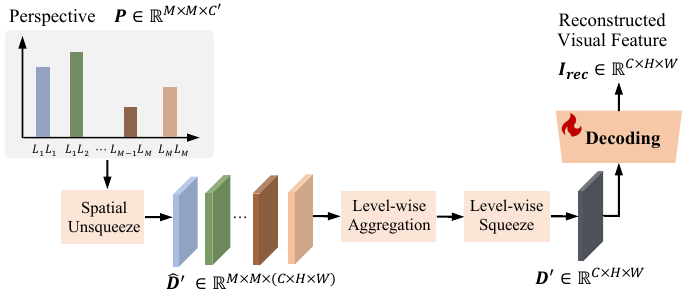}
  \caption{The illustration of Visual Reconstruction (VR). It is essentially the inverse of the PDD process, which reconstructs the visual feature from perspective representation.}
  \label{fig_vr}
\end{figure}

\paragraph{Perspective Prototype Extraction.} Besides $\mathbf{I}$, we utilize a pretrained SuperPoint \cite{superpoint} network to extract point-ness map $\mathbf{S}$ of the original image, which serve as the supporting structure of image perspective. The theoretical basis behind this motivation stems from related research such as multi-view visual understanding\cite{ma2021homography,dong2019fast,yaugmur2023improved}, which utilizes the structured spatial information of keypoints as support points to characterize the spatial structure of the image's perspective during actual photography  \cite{zeng2018rethinking,liu2023geometrized,zhang19963d,li2020srhen}. Next, we apply point-ness attention to $\mathbf{I}$ and $\mathbf{S}$ to obtain a basic perspective prototype,
\begin{equation}
    \mathbf{F} = {\rm Softmax}(\frac{\mathbf{I}\times \mathbf{S}^{\top}} {\sqrt{C}})\times \mathbf{S},
\end{equation}
\noindent where $\top$ is the matrix transposition operation. $\mathbf{F} \in \mathbb{R}^{C\times H \times W}$ is the perspective-structured visual feature.

\paragraph{Discretization.}  In order to effectively represent perspective within deep neural networks, we firstly perform Discretization on it. Inspired by \cite{stlnet,sstkd}, PDD defines $M$ levels by equally dividing $\mathbf{F}$ in the spatial dimension. Specifically the $m$-th ($m\in[1, M]$) level $\mathbf{L}_m \in \mathbb{R}^{C}$ is computed as:
\begin{equation}
    \mathbf{L}_m = \frac{m \cdot {\rm max}(\mathbf{F}) + (M-m)\cdot {\rm min}(\mathbf{F})}{M}.
\end{equation}
\noindent Next, based on $\mathbf{L}$, we use $M$ discretization functions to discretize  $F$ into 
an embedding $\mathbf{D}\in \mathbb{R}^{M \times C \times H \times W}$, where each $\mathbf{F}_{i,j}(i \in [1, H], j\in [1, W])$ is discretized into $\mathbf{D}_{i, j}\in \mathbb{R}^{M \times C}$. 
Specifically, 
\begin{equation}
    \mathbf{D}_{m, i, j} = \left\{
                                \begin{aligned}
                                 &\mathbf{e} - |\mathbf{L}_m - \mathbf{F}_{i,j}|  \\ 
                                &~~~~~~~~~~~~~~~~~~ {\rm if} ~~ |\mathbf{L}_m - \mathbf{F}_{i,j}| < \frac{\mathbf{L}_{m+1}-\mathbf{L}_{m}}{2} \\
                                 &\mathbf{0} ~~~~~~~~~~~~~~~~ {\rm else}
                                \end{aligned}
                                \right.,
\end{equation}
\noindent where $\mathbf{e} \in \mathbb{R}^{C}, \mathbf{0} \in \mathbb{R}^{C}$ denote the Unit vector and Zero vector respectively. Note a spireshaped function is used instead of a binary one to ensure that the operation is differentiable. 
$\mathbf{D}$ can be viewed as the form of a histogram for perspective distribution. According to the theory of photography, the perspective is related to spatial relationship between objects/pixels within the image. Correspondingly, we encode the spatial pixel co-occurring relationships by:
\begin{equation}
\begin{aligned}
    & \hat{\mathbf{D}}_{i,j} = \mathbf{D}_{i,j} \mathbf{D}_{i,j}^\top, ~~~ \hat{\mathbf{L}} = \{\hat{\mathbf{L}}_{\hat{m}}\}, \\
    \hat{m}&=[m_1, m_2],  m_1\in[1, M_1], m_2\in [1, M_2],
\end{aligned}
\end{equation}
\noindent where $M_1=M_2=M$, $\hat{\mathbf{q}} \in \mathbb{R}^{M \times M \times C \times H \times W}$ is spatial co-occurring perspective histogram, and $\hat{\mathbf{L}}$ is  the  corresponding co-occurring level. $\top$ indicates matrix transposition.

\paragraph{Statistical Representation.} Further, we construct the perspective graph $\mathbf{G}\in \mathbb{R}^{M\times M \times C'}$ by squeezing the spatial dimensions in form of counting the  intensity of each level, then encoding with position corresponding relationship \cite{stlnet,sstkd}:
\begin{equation}
\begin{aligned}
    \mathbf{G}_{\hat{m}} = & \Phi(\frac{\sum^{H}_{i=1}\sum^{W}_{j=1} \hat{\mathbf{D}}_{m, i, j}}{\sum^{M_1}_{m_1=1}\sum^{M_2}_{m_2=1}\sum^{H}_{i=1}\sum^{W}_{j=1} \hat{\mathbf{D}}_{m, n, i, j}}  \\
    & + \sum_{i=1}^{H} \sum_{j=1}^{W} \hat{\mathbf{D}}_{m_1, m_2, i, j}{\rm Pos\_Enc}(i, j)),
\end{aligned}  
\end{equation}
\noindent 
\noindent where ${\rm Pos\_Enc}$ indicates the position encoding operation. ${\rm Cat}$ is the concatenation operation. $\Phi$ is MLP layer for dimension normalization. As shown in Figure \ref{fig_pdd}, $\mathbf{G}$ can be viewed as a Perspective Distribution Graph, where the levels $\{L_1L_1, L_1L_2, ..., L_{M-1}L_{M}, L_ML_M\}$ are nodes. Finally, we correlate the perspective graph over all nodes to generate the overall perspective representation $\mathbf{P} \in \mathbb{R}^{M\times M \times C'}$:
\begin{equation}
    \mathbf{P} = \mathop{{\rm MA}}\limits_{\hat{m}}(\mathbf{G}_{\hat{m}}),
\end{equation}
\noindent where $\mathop{{\rm MA}}\limits_{\hat{m}}(\cdot)$ indicate performing multi-head attention over all $\hat{m}$ levels. By decomposing  in  form of discrete  graph  in latent embedding space, $\mathbf{P}$ is formulated as  pointness-structure-aware  thus able to represent the overall perspective information of the image.

\subsection{Visual Reconstruction} \label{sec_vr}

As illustrated in Section \ref{sec_overview} and Figure \ref{fig_vr}, we exploit the VR  to reconstruct visual feature $\mathbf{I}_{rec}$ from the perspective $\mathbf{P}$. It is essentially the inverse of the PDD process. Given the input perspective feature $\mathbf{P}$, we first restore the spatial dimension using an unsqueeze operation to obtain $\hat{\mathbf{D}}'\in \mathbb{R}^{M\times M\times (C\times H \times W)}$. Furthermore, we perform level-wise aggregation to mutually enhance the restored spatial information in form of multi-head attention. Afterward, we embed the discretization levels into the visual feature using a level-wise squeeze method to obtain $\mathbf{D}'\in \mathbb{R}^{C\times H \times W}$, and finally, a transformer decoder is used to reconstruct the final visual feature $\mathbf{I}_{rec}$. All the unsqueeze and squeeze processes can be accomplished using simple multi-layer MLPs.

In order to enable the reconstruction ability \cite{urur}, we use a reconstruction loss to supervise the decoding process to constrain the structure affinity between $\mathbf{I}$ and $\mathbf{I}_{rec}$,
\begin{equation}
\begin{aligned}
    \mathcal{L}_{rec} &= ||\mathbf{I}^{\top} \mathbf{I} - \mathbf{I}_{rec}^{\top} \mathbf{I}_{rec} ||_2.
\end{aligned}
\end{equation}

\begin{figure}
  \centering
  \includegraphics[width=0.5\linewidth]{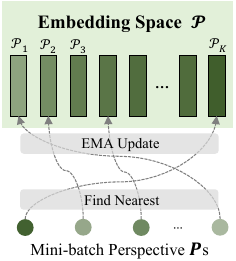}
  \caption{The illustration of the construction of Perspective Embedding Space $\mathcal{P}$, where $K$ is the size of $\mathcal{P}$ and $\mathcal{P}_k$ is the $k$-th embedding in $\mathcal{P}$ ($k\in [1,K]$). In training, we update $\mathcal{P}$ as a function of the exponential moving average (EMA) of the incoming mini-batch perspective $\mathbf{P}$s from PDD.}
  \label{fig_ema}
\end{figure}

\subsection{Perspective Embedding Space} \label{sec_ema}

Based on the decomposed perspective representation $\mathbf{P}$, we define a learnable latent perspective embedding space $\mathcal{P}\in \mathbb{R}^{K\times (M\times M \times C')} = \{\mathcal{P}_k \in \mathbb{R}^{M\times M\times C'}\} (k \in [1, K])$, where $K$ is the size of the latent space. As shown in Figure \ref{fig_ema}, in training, we update $\mathcal{P}$ as function of exponential moving average (EMA) of the coming mini-batch perspective representations $\mathbf{P}$s from PDD:
\begin{equation}
\begin{aligned}
& {\rm For~ each~~}  \mathbf{P} {\rm ~~ in~~ mini\-batch~~} \mathbf{P}{\rm s}, \\
    & \mathcal{P}_k := \left\{ 
    \begin{aligned}
    \alpha \mathcal{P}_k + (1 & -\alpha) \mathbf{P} \\ & {\rm if} ~~~ \mathop{\arg\min}\limits_{k}||\mathcal{P}_k - \mathbf{P}||_2 = k \\
    &\mathcal{P}_k ~~~~~~~~~~~~~~~~~~~~~~~~~~ {\rm else}
    \end{aligned}
    \right.,
\end{aligned}
\end{equation}
\noindent where $\alpha$ is the moving weight and we find $\alpha=0.9$ to work well in practice.

Then, we can obtain the overall perspective distribution of the whole dataset in form of Gaussian Mixed Model (GMM) over $\mathcal{P}$, as,
\begin{equation}
\begin{aligned}
    p = \sum_{k=1}^{K} \pi_k  \mathcal{N}(\mathcal{P}_k,  \mathbf{\Sigma}_k), ~~ \sum_{k=1}^K  \pi_k = 1, ~~ 0 \le \pi_k \le 1,
\end{aligned}
\label{gmm}
\end{equation}
\noindent where $\mathcal{N}(\cdot)$ indicates the Gaussian Distribution, the $k$-th component of GMM has the center of $\mathcal{P}^k$ with the variance of $\mathbf{\Sigma}_k$, and $\pi_k$ is the mixture coefficient.

\begin{figure}
  \centering
  \includegraphics[width=\linewidth]{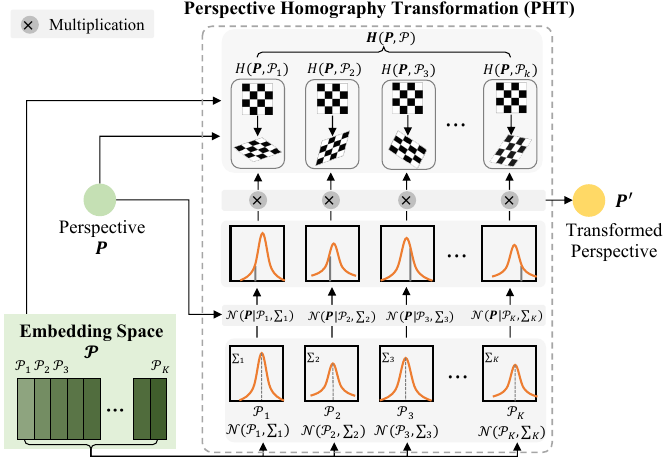}
  \caption{The illustration of PHT. Based on the characteristics of projection transformation, we transform the original perspective $\mathbf{P}$ to generate another descriptive perspective $\mathbf{P}'$, which corresponds to an alternative spatial projection of the image.}
  \label{fig_pht}
\end{figure}

\subsection{Perspective Homography Transformation}  \label{sec_pht}

In line with photography theory \cite{photo_theory}, we treat the perspective $\mathbf{P}$ as a spatial projection of the image \cite{malis2007deeper}. Based on the characteristics of projection transformation, we transform the perspective to generate another descriptive perspective $\mathbf{P}'$, which corresponds to an alternative spatial projection of the image. Based on Homography concept \cite{homography}, the original perspective $\mathbf{P}$ and the transformed perspective $\mathbf{P}'$ form a Homography, and this transformation can be realized with a Homography matrix.

As shown in Figure \ref{fig_pht}, in concrete, we generate a close semantic-related transformation $\mathbf{P}'$ of the original perspective $\mathbf{P}$, by leveraging the all the perspective distribution and performing Homography transformation:
\begin{equation}
\begin{aligned}
    \mathbf{P}' &= \mathbf{H} (\mathbf{P}, \mathcal{P}) \mathbf{P} = \sum_{k=1}^{K} H(\mathbf{P}, \mathcal{P}_k) p(\mathbf{P})  \\
    & = \sum_{k=1}^{K} \pi_n \mathcal{N}(\mathbf{P} | \mathcal{P}_k, \mathbf{\Sigma}_k) H(\mathbf{P}, \mathcal{P}_k), 
\end{aligned} 
\end{equation}

\noindent where $\mathbf{H}(\mathbf{P}, \mathcal{P})$ is the Homography Matrix of the original perspective $\mathbf{P}$ over all the perspective embedding space $\mathcal{P}$, and calculated by the sum of Homography Matrix of $\mathbf{P}$ over each $\mathcal{P}_k$ in the space, that is, $\sum_{k=1}^{K} H(\mathbf{P}, \mathcal{P}_k)$. $H(\mathbf{P}, \mathcal{P}_k)$ indicates the estimation of Homography matrix between $\mathbf{P}$ and $\mathcal{P}_k$. Mathematically, Homography matrix can represent the coordinates transformation from one perspective plane to another. In our formulation, $\mathbf{P}$ and $\mathcal{P}_k$ act as two perspective planes, thus $H(\mathbf{P}, \mathcal{P}_k)$ can characterize the perspective transformation from $\mathbf{P}$ to $\mathcal{P}_k$. 

Finally, we can obtain the corresponding visual feature $\mathbf{I}'_{rec}$ with the transformed perspective $\mathbf{P}'$, by applying the Visual Reconstruction (VR) module to $\mathbf{P}'$.

\begin{table}
\centering
\caption{The scenes, datasets and task settings in our experiments.}

\scalebox{0.68}{\begin{tabular}{c|c|c}
\toprule
\textbf{Scenes}       & \textbf{Semantic Segmentation}                    & \textbf{Object Detection} \\ \midrule
Auto-Driving & Cityscapes \cite{cityscapes}                               & -               \\ \midrule
UAV          & \begin{tabular}[c]{@{}l@{}} UDD6 \cite{udd} \\ iSAID \cite{isaid}\\ UAVid \cite{uavid} \\ Aeroscapes \cite{aeroscapes} \\ DroneSeg \end{tabular} & \begin{tabular}[c]{@{}l@{}} VisDrone \cite{visdrone} \\ UAVDT \cite{wu2021deep} \end{tabular} \\ \midrule
Daily Photos & ADE20K \cite{ade20k}                                   & -               \\ \bottomrule
\end{tabular}}
\label{dataset}
\end{table}

\begin{table}[]
\centering
\caption{UAV Scene Segmentation: Comparison with state-of-the-arts on UDD6, iSAID, UAVid, Aeroscapes and our proposed DroneSeg datasets.}
\scalebox{0.7}{\begin{tabular}{c|c|ccccc}
\toprule
\multirow{2}{*}{\textbf{Method}} &  \multirow{2}{*}{\textbf{Param.}}                       & \multicolumn{5}{c}{\textbf{mIoU} (\%)}                                    \\ \cmidrule{3-7} 
                                                        &       & \textbf{UDD6}  & \textbf{iSAID} & \textbf{UAVid} & \textbf{Aeroscapes} & \textbf{DroneSeg} \\ \midrule
Deeplab    &   63M           & 71.84 & 59.20                    & 56.82 & 51.40      & 38.69    \\
OCR\_W48      &        70M                    & 73.37 & 62.73                    & 63.10 & 58.19      & 43.10    \\
PSPNet        &  67M                                               & 72.95 & 60.30                    & 58.20  & 57.98      & 37.03    \\
FarSeg      &  -                   & -     & 63.70                    & -     & -          & -        \\
PFNet    &    -                                                   & -     & 66.90                    & -     & -          & -        \\
SCO             &  -                                 & -     & 69.10                    & -     & -          & -        \\ \midrule
SETR            &  318.3M                                              & 68.00 & 62.77                    & 58.52 & 50.34      & 48.23    \\
UperNet          &   234.0M                                            & 73.13 & 66.45                    & 61.91 & 64.32      & 53.34    \\
PoolFormer   &  77.1M                                                 & 74.54 & 65.55                    & 61.73 & 62.27      & 53.94    \\
SegFormer    &  84.7M                                                 & 74.28 & 67.19                    & 62.01 & 66.40      & 55.33    \\ \midrule
\begin{tabular}[c]{@{}c@{}}\textbf{DLPL} \\ (SETR)\end{tabular}   &  321.1M    & \textbf{72.62} & \textbf{65.45}                    & \textbf{60.23} & \textbf{55.02}      & \textbf{52.94}    \\ \midrule
\begin{tabular}[c]{@{}c@{}}\textbf{DLPL} \\ (UperNet)\end{tabular}   & 236.8M & \textbf{76.00} & \textbf{68.81}                    & \textbf{63.27} & \textbf{66.25}      & \textbf{56.10}    \\ \midrule
\begin{tabular}[c]{@{}c@{}}\textbf{DLPL} \\ (PoolFormer)\end{tabular}  &  79.9M   & \textbf{75.98} & \textbf{68.57}                    & \textbf{64.02} & \textbf{65.03}      & \textbf{56.44}    \\ \midrule
\begin{tabular}[c]{@{}c@{}}\textbf{DLPL} \\ (SegFormer)\end{tabular} &  87.5M & \textbf{76.88} & \textbf{70.60}                    & \textbf{65.30} & \textbf{68.22}      & \textbf{59.10}   \\ \bottomrule
\end{tabular}}

\label{exp_sota}
\end{table}

\begin{table}[tp]
\centering
\caption{UAV Scene Detection: Comparison with state-of-the-arts on VisDrone and UAVDT datasets.}
\scalebox{0.75}{\begin{tabular}{c|ccc|ccc}
\toprule
\multirow{2}{*}{\textbf{Method}} & \multicolumn{3}{c|}{\textbf{VisDrone}\textit{-dev}} & \multicolumn{3}{c}{\textbf{UAVDT}\textit{-test}} \\ \cmidrule{2-7} 
                        & \textbf{AP}      & \textbf{AP}$_{50}$  & \textbf{AP}$_{75}$  & \textbf{AP}     & \textbf{AP}$_{50}$ & \textbf{AP}$_{75}$ \\ \midrule
Dynamic R-CNN          & 13.70    & 24.70       & 13.50       & 57.60   & 65.30      & 64.6      \\
RetinaNet               & 18.94   & 31.67      & 20.25      & 33.95  & -         & -         \\
FRCNN(PVT2-B0)          & 19.40    & 31.80       & 20.80       & 60.80   & 71.50      & 69.8      \\
RefineDet               & 19.89   & 37.27      & 20.18      & -      & -         & -         \\
DetNet                  & 20.07   & 37.54      & 21.26      & -      & -         & -         \\
FPN                     & 22.06   & 39.57      & 22.50      & 49.05  & -         & -         \\
Light-RCNN              & 22.08   & 39.56      & 23.24      & -      & -         & -         \\
FRCNN(ResNet-50)        & 22.40    & 37.20       & 23.70       & 55.70   & 69.80      & 66.90      \\
CornerNet               & 23.43   & 41.18      & 25.02      & -      & -         & -         \\
Cascade-RCNN            & 25.20    & 40.10       & 26.80       & 61.50   & 73.50      & 71.70      \\ \midrule
DETR                    & 24.32   & 40.67      & 25.89      &  61.10      &    73.41       &    71.20       \\
Deformable DETR      &    26.28     &   43.02         &     27.90       &    62.99    &   76.32        &    73.78       \\
Sparse DETR     &    26.98     &  44.30    &  28.70      &   63.21   &     76.81      &    74.22      \\ \midrule
\begin{tabular}[c]{@{}c@{}}\textbf{DLPL} \\ (DETR)\end{tabular}             & \textbf{26.45}   & \textbf{43.28}      & \textbf{27.95}      &    \textbf{63.36}  &   \textbf{76.80}     &   \textbf{74.00}      \\ \midrule
\begin{tabular}[c]{@{}c@{}}\textbf{DLPL} \\ (Deformable DETR)\end{tabular}         &     \textbf{28.00}    &   \textbf{45.89}    &   \textbf{29.60}         &    \textbf{65.07 }  &    \textbf{78.67}       &    \textbf{76.01 }      \\ \midrule
\begin{tabular}[c]{@{}c@{}}\textbf{DLPL} \\ (Sparse DETR)\end{tabular}      &    \textbf{28.75}     &     \textbf{46.38}       &     \textbf{30.55 }      &    \textbf{65.57}    &     \textbf{79.02}      &      \textbf{76.92}     \\ \bottomrule
\end{tabular}}
\label{exp_sota_det}
\end{table}

\subsection{Perspective-Invariant Learning}  \label{sec_pia}

By the PHT and VR, we obtain the semantic-related perspective-transformed visual feature $\mathbf{I}'_{rec}$ for the original visual feature $\mathbf{I}$. As seen that $\mathbf{I}$ and $\mathbf{I}'_{rec}$ contain closely identical scene context and semantic information yet differ in perspective. Next, they are aggregated by the multi-level Perspective Invariant Attention (PIA)  to leverage the relationship between the $\mathbf{I}$ (with original perspective $\mathbf{P}$) and $\mathbf{I}_{rec}'$ (with transformed perspective $\mathbf{P}'$). Formally, PIA is formulated as a cross-attention manner \cite{crossatten}:
\begin{equation}
\begin{aligned}
    {\rm PIA}(\mathbf{I}, \mathbf{I}_{rec}') = {\rm Softmax}(\frac{\mathbf{I} \times \mathbf{I}_{rec}^{'\top}}{\sqrt{C}}) \times  \mathbf{I}_{rec}^{'}.
\end{aligned}
\end{equation}
The fundamental concept is to drive the model to jointly learn perspective-invariant semantic information in the image from multiple distinct viewpoints. Each layer of PIA generates a visual feature containing a new perspective. We cascade multiple layers of PIA and, through repeated perspective-invariant fusion, ultimately enhance the model's robustness to perspective in practical applications (e.g. auto-driving or UAV flights).

\subsection{Optimization}

The overall loss $\mathcal{L}$ is intuitive: the combination of the main task-specific loss $\mathcal{L}_{task}$  and the reconstruction loss $\mathcal{L}_{rec}$:
\begin{equation}
    \mathcal{L} = \mathcal{L}_{task} + \lambda \mathcal{L}_{rec},
    \label{loss}
\end{equation}
\noindent where $\lambda$ is the loss weight and set to $0.4$.

\section{Experiments}

\subsection{Scenes, Datasets, Tasks and Implementation Details}

 DLPL is a general and applicable to a broad range of scenes and tasks. In our experiments, we focus  on three types of scenarios (auto-driving, UAV, daily photos) and two fine-grained tasks (segmentation and detection). Furthermore, we concentrate on the more challenging task of segmentation, especially within the UAV scenario due to its rich variety of scene changes.  Specifically, given the current lack of a large-scale UAV segmentation benchmark dataset within the field, we propose DroneSeg\footnote{We release the proposed DroneSeg dataset at https://github. com/jankyee/DroneSeg.}, the largest-scale and semantically richest fine-grained annotated UAV segmentation dataset, to date. 
 
 In all experiments, we adopt the MMSegmentation \cite{mmsegmentation} and MMDetection \cite{chen2019mmdetection} toolboxes as codebases and follow the default basic configurations. In DLPL block, we employ the pretrained and SuperPoint parameters in \cite{superpoint} and froze them in PDD. In our experiments, we incorporate DLPL with various representative ViT segmentors and detectors.

 \begin{table}[]
\centering
\caption{Auto-Driving and Daily photos Segmentation: Comparison with state-of-the-arts on Cityscapes and ADE20K datasets.}
\scalebox{0.77}{\begin{tabular}{c|c|c|cc}
\toprule
\multirow{2}{*}{\textbf{Method}} & \multirow{2}{*}{\textbf{Backbone}} & \multirow{2}{*}{\textbf{Pram.}} & \multicolumn{2}{c}{\textbf{mIoU}(\%)} \\ \cline{4-5} 
                        &                           &                        & \textbf{Cityscapes} val  & \textbf{ADE20K} val \\ \midrule
FCN                     & ResNet-101                & 68.6M                  & 76.6            & 41.4       \\
EncNet                  & ResNet-101                & 55.1M                  & 76.9            & 44.7       \\
PSPNet                  & ResNet-101                & 68.1M                  & 78.5            & 44.4       \\
CCNet                   & ResNet-101                & 68.9M                  & 80.2            & 45.2       \\
DeeplabV3+              & ResNet-101                & 62.7M                  & 80.9            & 44.1       \\
OCRNet                  & HRNet-W48                 & 70.5M                  & 81.1            & 45.6       \\ \midrule
SETR                    & ViT-Large                 & 318.3M                 & 82.2            & 50.2       \\
Segmenter               & ViT-Large                 & 307.0M                 & 80.7            & 52.2       \\
SegFormer               & MiT-B4                    & 64.1M                  & 83.8            & 51.1       \\
SegFormer               & MiT-B5                    & 84.7M                  & 84.0            & 51.8       \\\midrule
\begin{tabular}[c]{@{}c@{}}\textbf{DLPL} \\ (SETR)\end{tabular}              & ViT-Large                 & 321.1M                 & \textbf{83.9}      & \textbf{53.1} \\\midrule
\begin{tabular}[c]{@{}c@{}}\textbf{DLPL} \\ (Segmenter)\end{tabular}         & ViT-Large                 & 309.8M                 & \textbf{82.3}      & \textbf{53.5} \\\midrule
\begin{tabular}[c]{@{}c@{}}\textbf{DLPL} \\ (SegFormer)\end{tabular}         & MiT-B4                    & 66.9M                  & \textbf{85.0}      & \textbf{53.5} \\\midrule
\begin{tabular}[c]{@{}c@{}}\textbf{DLPL} \\ (SegFormer)\end{tabular}         & MiT-B5                    & 87.5M                  & \textbf{85.1}      & \textbf{53.7} \\\bottomrule
\end{tabular}}
\label{sota_city}
\end{table}

\subsection{Comparison with State-of-the-Arts}

\subsubsection{UAV Scene}

Compared to the other two types of scenarios, UAVs present more challenging perspective understanding due to the rich variety of perspective changes caused by real-time movement and deviation during flight. Therefore, segmentation and detection in UAV scenarios become the focus of our experiments. As shown in Tables \ref{exp_sota} and \ref{exp_sota_det}, we apply DLPL to the currently advanced ViT networks and achieve significant performance improvements in segmentation and detection with less than 3M additional parameters. Specifically, on our newly proposed DroneSeg dataset, which presents the  level of challenge to date, we achieve the highest improvement of nearly 5\% over four types of plain ViT. In the UAV detection scenario, where objects are relatively sparse, we also achieve an improvement of nearly 3\% over various plain ViT detectors. This consistent performance enhancement directly confirms the effectiveness, generality, and robustness of our proposed DLPL.

\subsubsection{Auto-Driving and Daily Photos} 

These two scenes typically also encompass various shooting perspectives. To further validate the universality of DLPL, we conduct experimental verifications within these two types of scenes as well. Specifically, we focused primarily on the segmentation task whose performance is more evidently affected by perspective changes. As shown in Table \ref{sota_city}, ViT detectors integrated with DLPL achieve highest performance increases of 3\% compared to the plain ViT detectors.

\subsection{Ablation Study}

We perform ablation studies on semantic segmentation on DroneSeg \textit{test} set, SegFormer is used as baseline network.

\begin{table}[]
\centering
\caption{The comparison of DLPL  with other perspective-oriented learning methods (data augmentation). Perspective-Vertical and Perspective-Horizontal means adjusting perspectives in vertical and horizontal directions, and are implemented by the default ``torchvision.transforms" interfaces in PyTorch \cite{pytorch}. For each naive augmentation, we try multiple magnitudes and find the best performance, for fair comparisons.}
\scalebox{0.87}{\begin{tabular}{cc|c}
\toprule
\multicolumn{2}{c|}{Perspective-Oriented Learning Method}                                  & mIoU (\%) \\ \midrule
\multicolumn{2}{c|}{Baseline (SegFormer w/o data augmentation)}                                    & 52.03     \\ \midrule
\multicolumn{1}{c|}{\multirow{5}{*}{\begin{tabular}[c]{@{}c@{}}Naive \\ Augmentation\end{tabular}}} & + Random Rotate                 & 52.94     \\
\multicolumn{1}{c|}{}                                    & + Random Scale                  & 53.09     \\
\multicolumn{1}{c|}{}                                    & + Random Perspective-Vertical   & 53.56     \\
\multicolumn{1}{c|}{}                                    & + Random Perspective-Horizontal & 53.42     \\ \cmidrule{2-3} 
\multicolumn{1}{c|}{}                                    & + Random Combination            & 55.33     \\ \midrule
\multicolumn{2}{c|}{AutoAugmentation \cite{autoaugment}}                                                      & 55.80     \\ \midrule
\multicolumn{2}{c|}{DLPL (SegFormer w/o data augmentation)}                                        & 58.67     \\
\multicolumn{2}{c|}{DLPL (SegFormer + Random Combination)}                                 & 59.10     \\
\multicolumn{2}{c|}{DLPL (SegFormer + AutoAugmentation)}                                 & 59.09     \\\bottomrule
\end{tabular}}
\label{exp_compare_view}
\end{table}

\subsubsection{The Efficacy of DLPL}
As discussed in Section 2, as one type of perspective-oriented learning, we compare DLPL with the various ordinary perspective augmentation techniques. 
We notice that in all the scenes, the perspective changes are almost always related to the flying height and angles, which can be reflected on the scales, rotations, perspective-vertical and perspective-horizontal  augmentations. So, we use the combination of these data augmentation methods for the comparison with DLPL. Extensively, we also include the auto-augmentation techniques \cite{autoaugment} to further support our claims.  As shown in Table \ref{exp_compare_view}, DLPL shows more potentials than both naive and auto augmentations. We also show that DLPL is still adaptive to the ordinary data augmentation to obtain further improvements.

\subsubsection{The Significance of PDD} 
For the first time, PDD provides  heuristic inspiration for latent perspective representation. We demonstrate the significance of PDD by comparing with other ordinary perspective representation techniques: (1) directly forcing the perspective learning using an ordinary Transformer network, (2)  representing the perspective by directly using $\mathbf{F}$, followed by MLPs. The former lacks of in-depth characterization, and the latter may result in coarse-grained learning. As shown in Table \ref{abl_pdd}, PDD shows undisputed brilliant performance, confirming its rationality and effectiveness.

 \begin{table}[]
\centering
\caption{The significance of PDD by comparing it with ordinary perspective representation.}
\scalebox{0.9}{\begin{tabular}{c|c}
\toprule
Perspective Representation & mIoU (\%) \\ \midrule
 Ordinary Transformer Network                 & 57.07    \\
Directly using $\mathbf{F}$  & 55.91    \\
\midrule
PDD    & 59.10     \\ \bottomrule
\end{tabular}}
\label{abl_pdd}
\end{table}

\begin{table}[]
\centering
\caption{The effectiveness of PHT by comparing with ordinary perspective transformation methods.}
\scalebox{0.9}{\begin{tabular}{c|c}
\toprule
Perspective Transformation  & mIoU (\%) \\ \midrule
KNN                                                         & 56.57    \\
Momentum Update                                                   & 56.01    \\
Naive GMM                                                         & 56.94    \\ \midrule
PHT         & 59.10     \\ \bottomrule
\end{tabular}}
\label{exp_view_trans}
\end{table}

\subsubsection{Effectiveness of PHT}

PHT is for perspective transformation over the original perspective $\mathbf{P}$ and perspective embedding space $\mathcal{P}$, in the form of Homography.
In order to validate the effectiveness of such choice, we also implement some other perspective transformation methods for comparisons, including K-Nearest Neighbor (KNN), Momentum Update and naive GMM.  Results in Table \ref{exp_view_trans} demonstrate that PHT achieves the highest performance.

\subsubsection{Visualization}

\begin{figure}
\centering
  \includegraphics[width=1\linewidth]{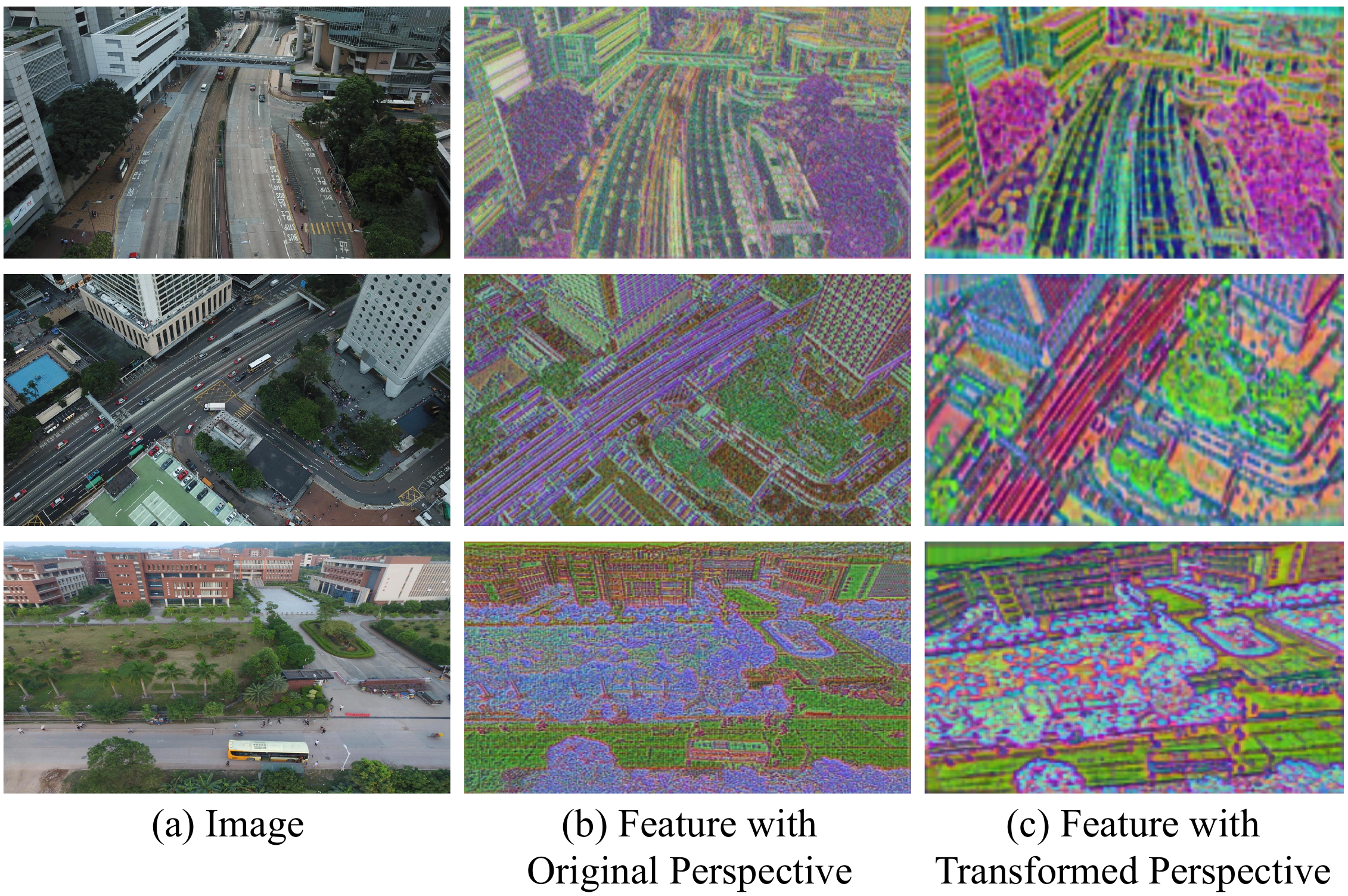}
  \caption{Visualization comparison on feature maps before and after DLPL. We  select three representative transformation intensities: weak (row 1), medium (row 2), and strong (row 3). We can find the overall views are obviously transformed after DLPL.}
  \label{feature_vis}
\end{figure}

To provide a clearer understanding of perspective transformation, we visualize and compare the feature maps before and after applying DLPL, as depicted in Figure \ref{feature_vis}. It shows that after passing through DLPL, the perspective reflected in visual feature maps are obviously transformed. 
For more in-depth insights, we  select representative transformation intensities: weak (row 1), medium (row 2), and strong (row 3). These intensities accurately capture the various stages of gradual perspective changes during network training, which is conducive to network training stability and also in line with the practical scene perspective changes such as UAV flights. As the training deepens, the network can eventually learn rich perspective diversity.

\section{Conclusion}

We introduce a Discrete Latent Perspective Learning (DLPL) framework that enhances Perspective-Invariant Learning using single-view images. The framework integrates Perspective Discrete Decomposition (PDD), Perspective Homography Transformation (PHT), and Multi-Perspective Fusion via Perspective Invariant Attention (PIA), overcoming the limitations of multi-view data collection and perspective augmentations. Extensive experiments have proven that DLPL can enhance the robustness of neural networks to perspective changes, achieving notable improvements in a wide range of scenes and tasks.

\section*{Acknowledgements}

This work was supported by the Anhui Provincial Natural Science Foundation under Grant 2108085UD12, the JKW Research Funds under Grant 20-163-14-LZ-001-004-01, the National Key R\&D Program of China under Grant 2020AAA0103902, NSFC (No. 62176155), Shanghai Municipal Science and Technology Major Project, China (2021SHZDZX0102).  

We acknowledge the support of GPU cluster built by MCC Lab of Information Science and Technology Institution, USTC. 

We would like to express our gratitude to Qi Zhu for his brilliant suggestions regarding the writings and figures of this paper, as well as the insightful comments from the Reviewers, Area Chairs, and Program Chairs.

\section*{Impact Statement}

This paper presents work whose goal is to advance the field of Machine Learning. There are many potential societal consequences of our work, none which we feel must be specifically highlighted here.

\nocite{langley00}

\bibliography{example_paper}

\begin{thebibliography}{62}
\providecommand{\natexlab}[1]{#1}
\providecommand{\url}[1]{\texttt{#1}}
\expandafter\ifx\csname urlstyle\endcsname\relax
  \providecommand{\doi}[1]{doi: #1}\else
  \providecommand{\doi}{doi: \begingroup \urlstyle{rm}\Url}\fi

\bibitem[LO1(1997)]{LO1997383}
Perspective-transformation-invariant generalized hough transform for perspective planar shape detection and matching.
\newblock \emph{Pattern Recognition}, 30\penalty0 (3):\penalty0 383--396, 1997.
\newblock ISSN 0031-3203.

\bibitem[YU2(2018)]{YU2018109}
A novel perspective invariant feature transform for rgb-d images.
\newblock \emph{Computer Vision and Image Understanding}, 167:\penalty0 109--120, 2018.
\newblock ISSN 1077-3142.

\bibitem[Bengio et~al.(2013)Bengio, Courville, and Vincent]{bengio2013representation}
Bengio, Y., Courville, A., and Vincent, P.
\newblock Representation learning: A review and new perspectives.
\newblock \emph{IEEE Transactions on Pattern Analysis and Machine Intelligence}, 35\penalty0 (8):\penalty0 1798--1828, 2013.

\bibitem[Carion et~al.(2020)Carion, Massa, Synnaeve, Usunier, Kirillov, and Zagoruyko]{detr}
Carion, N., Massa, F., Synnaeve, G., Usunier, N., Kirillov, A., and Zagoruyko, S.
\newblock End-to-end object detection with transformers.
\newblock In \emph{European Conference on Computer Vision}, pp.\  213--229, 2020.

\bibitem[Caron et~al.(2018)Caron, Bojanowski, Joulin, and Douze]{caron2018deep}
Caron, M., Bojanowski, P., Joulin, A., and Douze, M.
\newblock Deep clustering for unsupervised learning of visual features.
\newblock In \emph{Proceedings of the European conference on computer vision (ECCV)}, pp.\  132--149, 2018.

\bibitem[Caron et~al.(2020)Caron, Misra, Mairal, Goyal, Bojanowski, and Joulin]{caron2020unsupervised}
Caron, M., Misra, I., Mairal, J., Goyal, P., Bojanowski, P., and Joulin, A.
\newblock Unsupervised learning of visual features by contrasting cluster assignments.
\newblock \emph{Advances in neural information processing systems}, 33:\penalty0 9912--9924, 2020.

\bibitem[Chen et~al.(2021)Chen, Fan, and Panda]{crossatten}
Chen, C., Fan, Q., and Panda, R.
\newblock Crossvit: Cross-attention multi-scale vision transformer for image classification.
\newblock \emph{CoRR}, abs/2103.14899, 2021.

\bibitem[Chen et~al.(2019)Chen, Wang, Pang, Cao, Xiong, Li, Sun, Feng, Liu, Xu, et~al.]{chen2019mmdetection}
Chen, K., Wang, J., Pang, J., Cao, Y., Xiong, Y., Li, X., Sun, S., Feng, W., Liu, Z., Xu, J., et~al.
\newblock Mmdetection: Open mmlab detection toolbox and benchmark.
\newblock \emph{arXiv preprint arXiv:1906.07155}, 2019.

\bibitem[Chen et~al.(2018{\natexlab{a}})Chen, Zhu, Papandreou, Schroff, and Adam]{deeplabv3+}
Chen, L.-C., Zhu, Y., Papandreou, G., Schroff, F., and Adam, H.
\newblock Encoder-decoder with atrous separable convolution for semantic image segmentation.
\newblock In \emph{European Conference on Computer Vision}, pp.\  801--818, 2018{\natexlab{a}}.

\bibitem[Chen et~al.(2023{\natexlab{a}})Chen, Fu, Zhu, Mao, Zhang, Zang, and Sun]{chen2023deep3dsketch}
Chen, T., Fu, C., Zhu, L., Mao, P., Zhang, J., Zang, Y., and Sun, L.
\newblock Deep3dsketch: 3d modeling from free-hand sketches with view-and structural-aware adversarial training.
\newblock In \emph{ICASSP 2023-2023 IEEE International Conference on Acoustics, Speech and Signal Processing}, pp.\  1--5, 2023{\natexlab{a}}.

\bibitem[Chen et~al.(2023{\natexlab{b}})Chen, Zhu, Ding, Cao, Zhang, Wang, Li, Sun, Mao, and Zang]{chen2023sam}
Chen, T., Zhu, L., Ding, C., Cao, R., Zhang, S., Wang, Y., Li, Z., Sun, L., Mao, P., and Zang, Y.
\newblock Sam fails to segment anything?--sam-adapter: Adapting sam in underperformed scenes: Camouflage, shadow, and more.
\newblock \emph{arXiv preprint arXiv:2304.09148}, 2023{\natexlab{b}}.

\bibitem[Chen et~al.(2024{\natexlab{a}})Chen, Ding, Zhu, Zang, Liao, Li, and Sun]{chen2023reality3dsketch}
Chen, T., Ding, C., Zhu, L., Zang, Y., Liao, Y., Li, Z., and Sun, L.
\newblock Reality3dsketch: Rapid 3d modeling of objects from single freehand sketches.
\newblock \emph{IEEE Transactions on Multimedia}, 26:\penalty0 4859--4870, 2024{\natexlab{a}}.

\bibitem[Chen et~al.(2024{\natexlab{b}})Chen, Yu, Li, Zhang, Zhu, Ji, Zhang, Zang, Li, and Sun]{chen2024reasoning3d}
Chen, T., Yu, C., Li, J., Zhang, J., Zhu, L., Ji, D., Zhang, Y., Zang, Y., Li, Z., and Sun, L.
\newblock Reasoning3d -- grounding and reasoning in 3d: Fine-grained zero-shot open-vocabulary 3d reasoning part segmentation via large vision-language models.
\newblock \emph{arXiv preprint arXiv:2405.19326}, 2024{\natexlab{b}}.

\bibitem[Chen et~al.(2018{\natexlab{b}})Chen, Wang, Lu, Chen, and Wang]{udd}
Chen, Y., Wang, Y., Lu, P., Chen, Y., and Wang, G.
\newblock Large-scale structure from motion with semantic constraints of aerial images.
\newblock In \emph{Chinese Conference on Pattern Recognition and Computer Vision}, pp.\  347--359, 2018{\natexlab{b}}.

\bibitem[Cordts et~al.(2016)Cordts, Omran, Ramos, Rehfeld, Enzweiler, Benenson, Franke, Roth, and Schiele]{cityscapes}
Cordts, M., Omran, M., Ramos, S., Rehfeld, T., Enzweiler, M., Benenson, R., Franke, U., Roth, S., and Schiele, B.
\newblock The cityscapes dataset for semantic urban scene understanding.
\newblock In \emph{Proceedings of the IEEE conference on computer vision and pattern recognition}, pp.\  3213--3223, 2016.

\bibitem[Cubuk et~al.(2019)Cubuk, Zoph, Mane, Vasudevan, and Le]{autoaugment}
Cubuk, E.~D., Zoph, B., Mane, D., Vasudevan, V., and Le, Q.~V.
\newblock Autoaugment: Learning augmentation strategies from data.
\newblock In \emph{Proceedings of the IEEE/CVF Conference on Computer Vision and Pattern Recognition}, pp.\  113--123, 2019.

\bibitem[DeTone et~al.(2018)DeTone, Malisiewicz, and Rabinovich]{superpoint}
DeTone, D., Malisiewicz, T., and Rabinovich, A.
\newblock Superpoint: Self-supervised interest point detection and description.
\newblock In \emph{Proceedings of the IEEE Conference on Computer Vision and Pattern Recognition Workshops}, pp.\  224--236, 2018.

\bibitem[Dong et~al.(2019)Dong, Jiang, Huang, Bao, and Zhou]{dong2019fast}
Dong, J., Jiang, W., Huang, Q., Bao, H., and Zhou, X.
\newblock Fast and robust multi-person 3d pose estimation from multiple views.
\newblock In \emph{Proceedings of the IEEE/CVF Conference on Computer Vision and Pattern Recognition}, pp.\  7792--7801, 2019.

\bibitem[Elkins(2006)]{photo_theory}
Elkins, J.
\newblock Photography theory (art seminar).
\newblock Dec 2006.

\bibitem[Feng et~al.(2018)Feng, Ji, Wang, Chang, Ren, and Gan]{feng2018challenges}
Feng, W., Ji, D., Wang, Y., Chang, S., Ren, H., and Gan, W.
\newblock Challenges on large scale surveillance video analysis.
\newblock In \emph{IEEE Conference on Computer Vision and Pattern Recognition workshops}, pp.\  69--76, 2018.

\bibitem[Fu et~al.(2022)Fu, Zhang, Chen, Lu, Zhu, Zhou, Geiger, and Liao]{fu2022panoptic}
Fu, X., Zhang, S., Chen, T., Lu, Y., Zhu, L., Zhou, X., Geiger, A., and Liao, Y.
\newblock Panoptic nerf: 3d-to-2d label transfer for panoptic urban scene segmentation.
\newblock In \emph{2022 International Conference on 3D Vision}, pp.\  1--11, 2022.

\bibitem[Hu et~al.(2020)Hu, Ji, Gan, Bai, Wu, and Yan]{cdgc}
Hu, H., Ji, D., Gan, W., Bai, S., Wu, W., and Yan, J.
\newblock Class-wise dynamic graph convolution for semantic segmentation.
\newblock In \emph{European Conference on Computer Vision}, pp.\  1--17, 2020.

\bibitem[Ji et~al.(2019)Ji, Lu, and Zhang]{mscnn}
Ji, D., Lu, H., and Zhang, T.
\newblock End to end multi-scale convolutional neural network for crowd counting.
\newblock In \emph{Eleventh International Conference on Machine Vision}, volume 11041, pp.\  761--766, 2019.

\bibitem[Ji et~al.(2021)Ji, Wang, Hu, Gan, Wu, and Yan]{cagcn}
Ji, D., Wang, H., Hu, H., Gan, W., Wu, W., and Yan, J.
\newblock Context-aware graph convolution network for target re-identification.
\newblock In \emph{Proceedings of the AAAI Conference on Artificial Intelligence}, volume~35, pp.\  1646--1654, 2021.

\bibitem[Ji et~al.(2022)Ji, Wang, Tao, Huang, Hua, and Lu]{sstkd}
Ji, D., Wang, H., Tao, M., Huang, J., Hua, X.-S., and Lu, H.
\newblock Structural and statistical texture knowledge distillation for semantic segmentation.
\newblock In \emph{IEEE/CVF Conference on Computer Vision and Pattern Recognition}, pp.\  16876--16885, 2022.

\bibitem[Ji et~al.(2023{\natexlab{a}})Ji, Zhao, and Lu]{gpwformer}
Ji, D., Zhao, F., and Lu, H.
\newblock Guided patch-grouping wavelet transformer with spatial congruence for ultra-high resolution segmentation.
\newblock In \emph{Proceedings of the Thirty-Second International Joint Conference on Artificial Intelligence}, pp.\  920--928, 8 2023{\natexlab{a}}.

\bibitem[Ji et~al.(2023{\natexlab{b}})Ji, Zhao, Lu, Tao, and Ye]{urur}
Ji, D., Zhao, F., Lu, H., Tao, M., and Ye, J.
\newblock Ultra-high resolution segmentation with ultra-rich context: A novel benchmark.
\newblock In \emph{Proceedings of the IEEE/CVF Conference on Computer Vision and Pattern Recognition}, pp.\  23621--23630, June 2023{\natexlab{b}}.

\bibitem[Ji et~al.(2024{\natexlab{a}})Ji, Gao, Tao, Lu, and Zhao]{changenet}
Ji, D., Gao, S., Tao, M., Lu, H., and Zhao, F.
\newblock Changenet: Multi-temporal asymmetric change detection dataset.
\newblock In \emph{ICASSP 2024-2024 IEEE International Conference on Acoustics, Speech and Signal Processing}, pp.\  2725--2729, 2024{\natexlab{a}}.

\bibitem[Ji et~al.(2024{\natexlab{b}})Ji, Gao, Zhu, Zhu, Zhao, Xu, Lu, Ye, and Zhao]{homview}
Ji, D., Gao, S., Zhu, L., Zhu, Q., Zhao, Y., Xu, P., Lu, H., Ye, J., and Zhao, F.
\newblock View-centric multi-object tracking with homographic matching in moving uav.
\newblock \emph{arXiv preprint arXiv:2403.10830}, 2024{\natexlab{b}}.

\bibitem[Jing \& Tian(2020)Jing and Tian]{jing2020self}
Jing, L. and Tian, Y.
\newblock Self-supervised visual feature learning with deep neural networks: A survey.
\newblock \emph{IEEE transactions on pattern analysis and machine intelligence}, 43\penalty0 (11):\penalty0 4037--4058, 2020.

\bibitem[Kirillov et~al.(2023)Kirillov, Mintun, Ravi, Mao, Rolland, Gustafson, Xiao, Whitehead, Berg, Lo, et~al.]{sam}
Kirillov, A., Mintun, E., Ravi, N., Mao, H., Rolland, C., Gustafson, L., Xiao, T., Whitehead, S., Berg, A.~C., Lo, W.-Y., et~al.
\newblock Segment anything.
\newblock In \emph{Proceedings of the IEEE/CVF International Conference on Computer Vision}, pp.\  4015--4026, 2023.

\bibitem[Kriegman(2007)]{homography}
Kriegman, D.
\newblock Homography estimation.
\newblock \emph{Computer Vision I, CSE 252A, Winter}, 2007.

\bibitem[Li et~al.(2020)Li, Pei, and He]{li2020srhen}
Li, Y., Pei, W., and He, Z.
\newblock Srhen: stepwise-refining homography estimation network via parsing geometric correspondences in deep latent space.
\newblock In \emph{Proceedings of the 28th ACM International Conference on Multimedia}, pp.\  3063--3071, 2020.

\bibitem[Liu \& Li(2023)Liu and Li]{liu2023geometrized}
Liu, J. and Li, X.
\newblock Geometrized transformer for self-supervised homography estimation.
\newblock In \emph{Proceedings of the IEEE/CVF International Conference on Computer Vision}, pp.\  9556--9565, 2023.

\bibitem[Lyu et~al.(2020)Lyu, Vosselman, Xia, Yilmaz, and Yang]{uavid}
Lyu, Y., Vosselman, G., Xia, G.-S., Yilmaz, A., and Yang, M.~Y.
\newblock Uavid: A semantic segmentation dataset for uav imagery.
\newblock \emph{ISPRS Journal of Photogrammetry and Remote Sensing}, 165:\penalty0 108 -- 119, 2020.

\bibitem[Ma et~al.(2021)Ma, Chen, Liu, and Zhang]{ma2021homography}
Ma, L., Chen, K., Liu, J., and Zhang, J.
\newblock Homography-driven plane feature matching and pose estimation.
\newblock In \emph{2021 6th IEEE International Conference on Advanced Robotics and Mechatronics}, pp.\  791--796, 2021.

\bibitem[Malis \& Vargas~Villanueva(2007)Malis and Vargas~Villanueva]{malis2007deeper}
Malis, E. and Vargas~Villanueva, M.
\newblock Deeper understanding of the homography decomposition for vision-based control.
\newblock 2007.

\bibitem[MMSegmentation(2020)]{mmsegmentation}
MMSegmentation.
\newblock Mmsegmentation: Openmmlab semantic segmentation toolbox and benchmark.
\newblock \emph{Availabe online: https://github. com/open-mmlab/mmsegmentation}, 2020.

\bibitem[Mumuni \& Mumuni(2022)Mumuni and Mumuni]{MUMUNI2022100258}
Mumuni, A. and Mumuni, F.
\newblock Data augmentation: A comprehensive survey of modern approaches.
\newblock \emph{Array}, 16:\penalty0 100258, 2022.
\newblock ISSN 2590-0056.

\bibitem[Nigam et~al.(2018)Nigam, Huang, and Ramanan]{aeroscapes}
Nigam, I., Huang, C., and Ramanan, D.
\newblock Ensemble knowledge transfer for semantic segmentation.
\newblock In \emph{2018 IEEE Winter Conference on Applications of Computer Vision}, pp.\  1499--1508, 2018.

\bibitem[Paszke et~al.(2019)Paszke, Gross, Massa, Lerer, Bradbury, Chanan, Killeen, Lin, Gimelshein, Antiga, et~al.]{pytorch}
Paszke, A., Gross, S., Massa, F., Lerer, A., Bradbury, J., Chanan, G., Killeen, T., Lin, Z., Gimelshein, N., Antiga, L., et~al.
\newblock Pytorch: An imperative style, high-performance deep learning library.
\newblock \emph{Advances in Neural Information Processing Systems}, 32, 2019.

\bibitem[Wang et~al.(2021{\natexlab{a}})Wang, Jiao, Liu, Li, Liu, Ji, and Gan]{ipgn}
Wang, H., Jiao, L., Liu, F., Li, L., Liu, X., Ji, D., and Gan, W.
\newblock Ipgn: Interactiveness proposal graph network for human-object interaction detection.
\newblock \emph{IEEE Transactions on Image Processing}, 30:\penalty0 6583--6593, 2021{\natexlab{a}}.

\bibitem[Wang et~al.(2021{\natexlab{b}})Wang, Jiao, Liu, Li, Liu, Ji, and Gan]{wang2021learning}
Wang, H., Jiao, L., Liu, F., Li, L., Liu, X., Ji, D., and Gan, W.
\newblock Learning social spatio-temporal relation graph in the wild and a video benchmark.
\newblock \emph{IEEE Transactions on Neural Networks and Learning Systems}, 34\penalty0 (6):\penalty0 2951--2964, 2021{\natexlab{b}}.

\bibitem[Wang et~al.(2023)Wang, Cheng, Chen, Shao, Zhu, Wu, Liu, and Zhu]{wang2023fvp}
Wang, Y., Cheng, J., Chen, Y., Shao, S., Zhu, L., Wu, Z., Liu, T., and Zhu, H.
\newblock Fvp: Fourier visual prompting for source-free unsupervised domain adaptation of medical image segmentation.
\newblock \emph{IEEE Transactions on Medical Imaging}, 2023.

\bibitem[Waqas~Zamir et~al.(2019)Waqas~Zamir, Arora, Gupta, Khan, Sun, Shahbaz~Khan, Zhu, Shao, Xia, and Bai]{isaid}
Waqas~Zamir, S., Arora, A., Gupta, A., Khan, S., Sun, G., Shahbaz~Khan, F., Zhu, F., Shao, L., Xia, G.-S., and Bai, X.
\newblock isaid: A large-scale dataset for instance segmentation in aerial images.
\newblock In \emph{Proceedings of the IEEE Conference on Computer Vision and Pattern Recognition Workshops}, pp.\  28--37, 2019.

\bibitem[Wu et~al.(2021)Wu, Li, Hong, Tao, and Du]{wu2021deep}
Wu, X., Li, W., Hong, D., Tao, R., and Du, Q.
\newblock Deep learning for unmanned aerial vehicle-based object detection and tracking: A survey.
\newblock \emph{IEEE Geoscience and Remote Sensing Magazine}, 10\penalty0 (1):\penalty0 91--124, 2021.

\bibitem[Xie et~al.(2021)Xie, Wang, Yu, Anandkumar, Alvarez, and Luo]{segformer}
Xie, E., Wang, W., Yu, Z., Anandkumar, A., Alvarez, J.~M., and Luo, P.
\newblock Segformer: Simple and efficient design for semantic segmentation with transformers.
\newblock \emph{Advances in neural information processing systems}, 34:\penalty0 12077--12090, 2021.

\bibitem[Ya{\u{g}}mur \& Ates(2023)Ya{\u{g}}mur and Ates]{yaugmur2023improved}
Ya{\u{g}}mur, {\.I}.~C. and Ates, H.~F.
\newblock Improved homographic adaptation for keypoint generation in cross-spectral registration of thermal and optical imagery.
\newblock In \emph{Image and Signal Processing for Remote Sensing XXIX}, volume 12733, pp.\  67--73, 2023.

\bibitem[Yu et~al.(2022)Yu, Luo, Zhou, Si, Zhou, Wang, Feng, and Yan]{poolformer}
Yu, W., Luo, M., Zhou, P., Si, C., Zhou, Y., Wang, X., Feng, J., and Yan, S.
\newblock Metaformer is actually what you need for vision.
\newblock In \emph{Proceedings of the IEEE/CVF Conference on Computer Vision and Pattern Recognition}, pp.\  10819--10829, 2022.

\bibitem[Yu et~al.(2023)Yu, Xu, Zhang, Liu, Ye, Wu, Yan, Zhu, Xiong, Liang, Chen, Cui, and Han]{mvimgnet}
Yu, X., Xu, M., Zhang, Y., Liu, H., Ye, C., Wu, Y., Yan, Z., Zhu, C., Xiong, Z., Liang, T., Chen, G., Cui, S., and Han, X.
\newblock Mvimgnet: A large-scale dataset of multi-view images.
\newblock In \emph{Proceedings of the IEEE/CVF Conference on Computer Vision and Pattern Recognition}, pp.\  9150--9161, June 2023.

\bibitem[Yuan et~al.(2020)Yuan, Chen, and Wang]{ocrnet}
Yuan, Y., Chen, X., and Wang, J.
\newblock Object-contextual representations for semantic segmentation.
\newblock In \emph{European Conference on Computer Vision}, pp.\  173--190, 2020.

\bibitem[Zeng et~al.(2018)Zeng, Denman, Sridharan, and Fookes]{zeng2018rethinking}
Zeng, R., Denman, S., Sridharan, S., and Fookes, C.
\newblock Rethinking planar homography estimation using perspective fields.
\newblock In \emph{Asian Conference on Computer Vision}, pp.\  571--586, 2018.

\bibitem[Zhang \& Hanson(1996)Zhang and Hanson]{zhang19963d}
Zhang, Z. and Hanson, A.~R.
\newblock 3d reconstruction based on homography mapping.
\newblock \emph{Proc. ARPA96}, pp.\  1007--1012, 1996.

\bibitem[Zheng et~al.(2021)Zheng, Lu, Zhao, Zhu, Luo, Wang, Fu, Feng, Xiang, Torr, and Zhang]{setr}
Zheng, S., Lu, J., Zhao, H., Zhu, X., Luo, Z., Wang, Y., Fu, Y., Feng, J., Xiang, T., Torr, P.~H., and Zhang, L.
\newblock Rethinking semantic segmentation from a sequence-to-sequence perspective with transformers.
\newblock In \emph{Proceedings of the IEEE/CVF Conference on Computer Vision and Pattern Recognition}, 2021.

\bibitem[Zhou et~al.(2017)Zhou, Zhao, Puig, Fidler, Barriuso, and Torralba]{ade20k}
Zhou, B., Zhao, H., Puig, X., Fidler, S., Barriuso, A., and Torralba, A.
\newblock Scene parsing through ade20k dataset.
\newblock In \emph{Proceedings of the IEEE Conference on Computer Vision and Pattern Recognition}, pp.\  633--641, 2017.

\bibitem[Zhu et~al.(2021{\natexlab{a}})Zhu, Ji, Zhu, Gan, Wu, and Yan]{stlnet}
Zhu, L., Ji, D., Zhu, S., Gan, W., Wu, W., and Yan, J.
\newblock Learning statistical texture for semantic segmentation.
\newblock In \emph{Proceedings of the IEEE/CVF Conference on Computer Vision and Pattern Recognition}, pp.\  12537--12546, June 2021{\natexlab{a}}.

\bibitem[Zhu et~al.(2023{\natexlab{a}})Zhu, Chen, Ji, Ye, and Liu]{llafs}
Zhu, L., Chen, T., Ji, D., Ye, J., and Liu, J.
\newblock Llafs: When large-language models meet few-shot segmentation.
\newblock \emph{arXiv preprint arXiv:2311.16926}, 2023{\natexlab{a}}.

\bibitem[Zhu et~al.(2023{\natexlab{b}})Zhu, Chen, Yin, See, and Liu]{zhu2023continual}
Zhu, L., Chen, T., Yin, J., See, S., and Liu, J.
\newblock Continual semantic segmentation with automatic memory sample selection.
\newblock In \emph{IEEE/CVF Conference on Computer Vision and Pattern Recognition}, pp.\  3082--3092, 2023{\natexlab{b}}.

\bibitem[Zhu et~al.(2023{\natexlab{c}})Zhu, Chen, Yin, See, and Liu]{zhu2023gabor}
Zhu, L., Chen, T., Yin, J., See, S., and Liu, J.
\newblock Learning gabor texture features for fine-grained recognition.
\newblock In \emph{IEEE/CVF International Conference on Computer Vision}, pp.\  1621--1631, 2023{\natexlab{c}}.

\bibitem[Zhu et~al.(2024{\natexlab{a}})Zhu, Chen, Yin, See, and Liu]{zhu2023add}
Zhu, L., Chen, T., Yin, J., See, S., and Liu, J.
\newblock Addressing background context bias in few-shot segmentation through iterative modulation.
\newblock In \emph{IEEE/CVF International Conference on Computer Vision}, 2024{\natexlab{a}}.

\bibitem[Zhu et~al.(2024{\natexlab{b}})Zhu, Ji, Chen, Xu, Ye, and Liu]{ibdnet}
Zhu, L., Ji, D., Chen, T., Xu, P., Ye, J., and Liu, J.
\newblock Ibd: Alleviating hallucinations in large vision-language models via image-biased decoding.
\newblock \emph{arXiv preprint arXiv:2402.18476}, 2024{\natexlab{b}}.

\bibitem[Zhu et~al.(2021{\natexlab{b}})Zhu, Wen, Du, Bian, Fan, Hu, and Ling]{visdrone}
Zhu, P., Wen, L., Du, D., Bian, X., Fan, H., Hu, Q., and Ling, H.
\newblock Detection and tracking meet drones challenge.
\newblock \emph{IEEE Transactions on Pattern Analysis and Machine Intelligence}, 44\penalty0 (11):\penalty0 7380--7399, 2021{\natexlab{b}}.

\end{thebibliography}
\bibliographystyle{icml2024}

\end{document}